\documentclass[letterpaper, 10 pt, conference]{ieeeconf}
\usepackage{graphicx} % Required for inserting images
\usepackage{amsmath}
\usepackage{amssymb}
\usepackage{booktabs}
\usepackage{array}
\usepackage{float}

\IEEEoverridecommandlockouts 
\title{\LARGE \bf
Gait Asymmetry from Unilateral Weakness and Improvement With Ankle Assistance: a Reinforcement Learning based Simulation Study
}

\author{Yifei Yuan, Ghaith Androwis, and Xianlian Zhou% <-this % stops a space
\thanks{Yifei Yuan and Xianlian Zhou are with the Department of Biomedical Engineering, New Jersey Institute of Technology, Newark, NJ 07102, USA (yy72@njit.edu, alexzhou@njit.edu). Ghaith Androwis (GAndrowis@kesslerfoundation.org) is with Kessler Foundation, West Orange, NJ, 07052, USA}
}

\let\oldthebibliography\thebibliography
\renewcommand{\thebibliography}[1]{%
  \oldthebibliography{#1}%
  \setlength{\itemsep}{0pt}%
  \setlength{\parskip}{0pt}%
  \setlength{\parsep}{0pt}%
}

\begin{document}

\maketitle
\thispagestyle{empty}
\pagestyle{empty}

% \linenumbers
% \setlength\linenumbersep{1pt}
\begin{abstract}
Unilateral muscle weakness often leads to asymmetric gait, disrupting interlimb coordination and stance timing. This study presents a reinforcement learning (RL)–based musculoskeletal simulation framework to (1) quantify how progressive unilateral muscle weakness affects gait symmetry and (2) evaluate whether ankle exoskeleton assistance can improve gait symmetry under impaired conditions. The overarching goal is to establish a simulation- and learning-based workflow that supports early controller development prior to patient experiments. Asymmetric gait was induced by reducing right-leg muscle strength to 75\%, 50\%, and 25\% of baseline. Gait asymmetry was quantified using toe-off timing, peak contact forces, and joint-level symmetry metrics. Increasing weakness produced progressively larger temporal and kinematic asymmetry, most pronounced at the ankle. Ankle range of motion symmetry degraded from near-symmetric behavior at 100\% strength (symmetry index, SI $=+6.4\%$; correlation $r=0.974$) to severe asymmetry at 25\% strength (SI $=-47.1\%$, $r=0.889$), accompanied by a load shift toward the unimpaired limb. At 50\% strength, ankle exoskeleton assistance improved kinematic symmetry relative to the unassisted impaired condition, reducing the magnitude of ankle SI from $25.8\%$ to $18.5\%$ and increasing ankle correlation from $r=0.948$ to $0.966$, although peak loading remained biased toward the unimpaired side. Overall, this framework supports controlled evaluation of impairment severity and assistive strategies, and provides a basis for future validation in human experiments.

\end{abstract}

\section{Introduction}
Muscle weakness in a single lower limb is common in conditions such as stroke, cerebral palsy, orthopedic injury, and age-related sarcopenia \cite{ryan2017sarcopenia}. Unilateral weakness often leads to gait asymmetry, including unequal stance time, uneven limb loading, and disrupted coordination between the legs \cite{sadeghi2000symmetry}. Such asymmetry is linked to higher metabolic cost, increased fall risk, and overuse of the stronger leg, which can further elevate injury risk over time \cite{ helme2021does}. For these reasons, improving gait symmetry has received increasing attention in rehabilitation science and assistive robotics, especially for individuals who rely on walking for daily mobility and independence.
 
Wearable exoskeletons have been developed to assist gait in individuals with weakness or motor impairment \cite{rodriguez2021systematic}. Many existing exoskeleton systems use predefined torque profiles, rule-based control strategies, or require manual tuning by users \cite{madinei2022novel, fricke2020automatic}. These approaches can be effective in certain cases, but they often struggle to adapt to different levels of impairment or to the individual’s compensatory strategies. Moreover, most prior work focuses on providing assistance or reducing effort rather than explicitly restoring gait symmetry \cite{kim2019potential}.

Musculoskeletal simulation provides a safe and effective way to study impaired gait without risk to human participants and allows a precise controlled manipulation of unilateral muscle weakness \cite{falisse2020physics}. Prior simulation studies have examined changes in kinematics or energy expenditure under weakness, but few have systematically evaluated how progressive unilateral muscle weakness affects stance-phase symmetry and inter-limb coordination \cite{van2012predictive}. Many existing models also rely on hand-crafted control rules or assume largely symmetric neuromuscular control architectures, which may limit their ability to capture how asymmetry naturally emerges when one limb is impaired.

Reinforcement learning (RL) offers a way to develop adaptive exoskeleton control policies that learn from interaction rather than relying on predefined control rules \cite{luo2023robust,ratnakumar2024predicting,luo2024experiment}. When combined with simulation, RL allows systematic testing of assistance strategies prior to human experiments and makes it possible to target explicit gait objectives such as symmetry \cite{wen2020wearer}. Importantly, an RL-based framework can model the coupled human–exoskeleton dynamics, allowing both to adapt during training instead of assuming a fixed or passive human response to assistance.

In this study, we utilize the MyoAssist reinforcement learning framework to model unilateral muscle weakness and the resulting gait asymmetry, and to evaluate ankle exoskeleton assistance strategies for improving gait symmetry. We simulate progressively increasing unilateral weakness in the right leg and quantify changes in stance time symmetry and joint coordination. We then train an RL controller to provide ankle assistance under impaired conditions. Our results show that unilateral weakness produces clear gait asymmetry and stance imbalance, and that RL-based ankle assistance can partially restore gait symmetry under the $50\%$ strength condition.
 
\section{Method}
\subsection{Reinforcement Learning Control Framework}

We used the MyoAssist reinforcement learning framework to couple a musculoskeletal human model with a wearable ankle exoskeleton device \cite{tan2025myoassist}. The human model included full three-dimensional (3D) body dynamics and 26 lower-limb muscles,  that generated joint torques through muscle activation. The framework supports a flexible multi-agent architecture in which a human policy produces muscle activations and an exoskeleton policy generates assistive joint commands. In this study, the exoskeleton policy was configured to provide assistive torques at the right ankle joint. Both agents share a 65-D observation; the exoskeleton outputs one normalized right-ankle command (ctrlrange $[-1,0]$) mapped to a peak torque of 100~N$\cdot$m. The command was mapped to a unidirectional plantarflexion assistance torque (negative sign in Fig.~\ref{fig:exo_torque_50}), with zero corresponding to no assistance.

Training was performed in two stages. In the first stage, we trained a baseline muscle control policy for the human model under unimpaired conditions (100\% strength, $\alpha=1.0$). In the second stage, we initialized transfer learning from the normal-walking baseline to multiple levels of unilateral muscle weakness (e.g., $\alpha \in \{0.75, 0.50, 0.25\}$), resulting in stable but asymmetric gait patterns under impaired conditions. We then jointly optimized the human and exoskeleton policies in a multi-agent setting \cite{baltes2025cooperative}. The exoskeleton policy learned to provide right-ankle assistance, while the human policy continued to adapt in response to the presence of the assistive torque.

We formulate the joint learning objective as
\begin{equation}
\label{eq:joint_obj}
(\pi_h^*, \pi_e^*) = \arg\max_{\pi_h,\pi_e}
\mathbb{E}\left[
\sum_{t=0}^{T}
\gamma^{t}\big(
r_t^{(h)} + \beta r_t^{(e)}
\big)
\right].
\end{equation}

where $\pi_h$ is the neuromuscular control policy for the 26-muscle human model and $\pi_e$ is the exoskeleton assistance policy. Equation~\ref{eq:joint_obj} gives the general multi-agent objective. In this study, both agents were optimized using the same imitation-style locomotion reward (Section~\ref{sec:reward}), i.e., $r_t^{(h)}=r_t^{(e)}=r_t$ and $\beta=1$. Symmetry was not included as an explicit reward term and was evaluated as an outcome of the learned assistance. In Stage~2, the two policies were updated concurrently using a shared centralized critic, while each policy maintained its own actor.

\begin{figure}[t]
    \centering
    \includegraphics[width=1.0\linewidth]{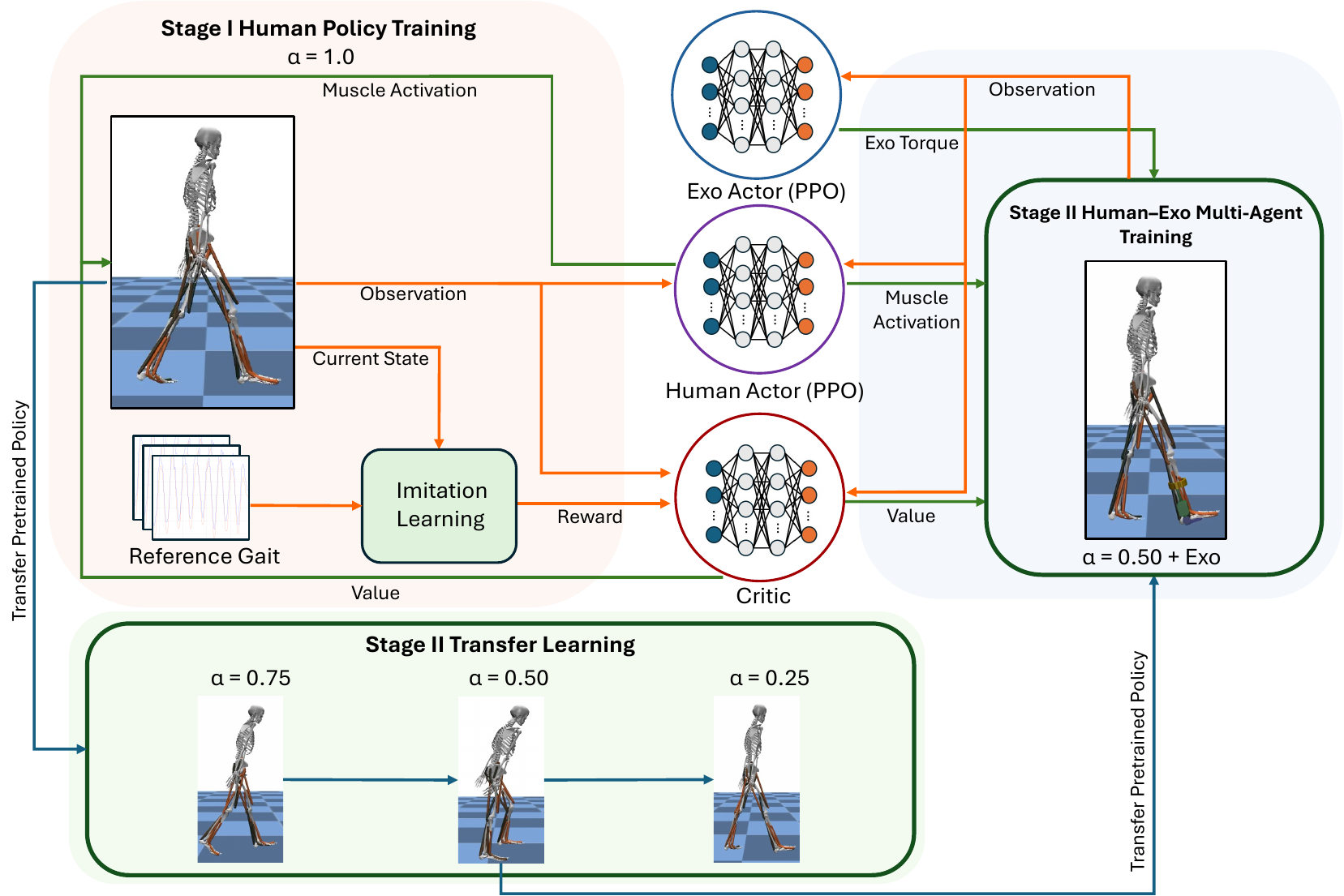}
    \caption{
    Reinforcement learning framework for human--exoskeleton multi-agent training, adapted from \cite{tan2025myoassist}. 
    Stage~1 trains a baseline human locomotion policy for normal walking. 
    Stage~2 performs transfer learning under progressive unilateral weakness; joint human--exoskeleton training for right-ankle assistance was performed only at $\alpha=0.5$ using PPO \cite{schulman2017proximal}.
    }
    \label{fig:framework}
\end{figure}

\subsection{Modeling Unilateral Muscle Weakness}

To study gait asymmetry arising from unilateral weakness, we systematically reduced the strength of muscles on the right leg only. Weakness was implemented by scaling the maximum isometric muscle force of each affected muscle as

\begin{equation}
F_{\text{max,weak}} = \alpha\, F_{\text{max,baseline}},
\end{equation}

where $\alpha$ is the strength scaling factor \cite{kainz2018influence}. We tested four levels of unilateral weakness,

\begin{equation}
\alpha \in \{1.00,\ 0.75,\ 0.50,\ 0.25\}.
\end{equation}
For clarity, these values correspond to 100\%, 75\%, 50\%, and 25\% of baseline muscle strength, respectively.

Here unilateral weakness was applied to the right leg only, while the left leg remained at full strength (100\% muscle strength). This setup isolates the impairment to one limb while allowing compensatory coordination to emerge naturally through the coupled multi-joint dynamics. For each weakness level, the human policy was fine-tuned from the Stage~1 baseline; human--exoskeleton joint training was only performed at $\alpha=0.5$ (50\% strength).

\subsection{Reward Structure}\label{sec:reward}

The controllers were trained with an imitation-style reward that tracks a reference gait trajectory while encouraging smooth muscle activations \cite{simos2025reinforcement}. The per-timestep reward is
\begin{equation}
r
=
w_{\text{qpos}}\, r_{\text{qpos}}
+
w_{\text{qvel}}\, r_{\text{qvel}}
+
w_{\text{ee}}\, r_{\text{ee}}
+
w_{\Delta a}\, r_{\Delta a}.
\label{eq:reward_total}
\end{equation}

The reward terms are computed as
\begin{subequations}
\label{eq:reward_terms}
\begin{align}
r_{\text{qpos}} &= \Delta t \sum_{i\in\mathcal{J}} w_i
\exp\!\left(-8\,(q_i-q_i^{\text{ref}})^2\right), \\
r_{\text{qvel}} &= \Delta t \sum_{i\in\mathcal{J}} w_i
\exp\!\left(-8\,(\dot q_i-\rho\,\dot q_i^{\text{ref}})^2\right), \\
r_{\text{ee}} &= \Delta t \cdot \frac{1}{|\mathcal{A}|}\sum_{k\in\mathcal{A}}
\exp\!\left(-5\,\|p_k-p_k^{\text{ref}}\|^2\right), \\
r_{\Delta a} &= \Delta t \cdot \frac{1}{N_m}\sum_{j=1}^{N_m}
\exp\!\left(-4\,(a_{j,t}-a_{j,t-1})^2\right),
\end{align}
\end{subequations}
where $\mathcal{J}$ denotes the set of tracked joints with weights $w_i$, $\mathcal{A}$ denotes the set of end-effectors (feet and toes), $N_m$ is the number of muscles, and $\Delta t$ is the control time step.
Here $q_i$ and $\dot q_i$ are joint positions and velocities, $p_k$ is the end-effector position, and $a_{j,t}$ is the activation of muscle $j$ at time $t$.
Superscript ``ref'' denotes the reference gait trajectory, and $\rho$ scales the reference joint velocities.

\subsection{Gait Symmetry Metrics and Analysis}\label{sec:metrics}

Gait symmetry was evaluated using the following temporal and kinematic measures:

\textbf{Stance-time asymmetry.}
Temporal asymmetry was quantified using the stance-time asymmetry metric \cite{herzog1989asymmetries},

\begin{equation}
\label{eq:stance_asym}
\text{Asym}_{\text{stance}}
=
\frac{T_{\text{left}} - T_{\text{right}}}
{\tfrac{1}{2}\left(T_{\text{left}} + T_{\text{right}}\right)},
\end{equation}

where $T_{\text{left}}$ and $T_{\text{right}}$ denote the stance durations of the left and right legs within a gait cycle. A value of zero indicates perfect symmetry, while the positive sign indicates the limb with longer stance.

\textbf{Joint-level symmetry.}
Kinematic symmetry was quantified using stance-phase joint-angle trajectories. For each gait cycle, the hip, knee, and ankle joint angles were segmented over the stance phase and time-normalized to a common phase variable $\phi \in [0,1]$, where $\phi=0$ corresponds to heel strike and $\phi=1$ corresponds to toe-off of the same limb. For each joint $j$, stance-phase range of motion (ROM) was computed separately for the left and right legs, and symmetry was summarized using the ROM-based symmetry index (SI) defined as
\begin{equation}
\label{eq:SI}
\text{SI}
=
\frac{ROM_{\text{left}} - ROM_{\text{right}}}
{\tfrac{1}{2}\left(ROM_{\text{left}} + ROM_{\text{right}}\right)}.
\end{equation}
Values of SI closer to zero indicate greater kinematic symmetry; a positive SI indicates larger ROM on the left limb, and a negative SI indicates larger ROM on the right limb.

\textbf{Trajectory correlation.}
To quantify inter-limb coordination in joint kinematics, we computed the Pearson correlation coefficient $r$ between the left and right joint angle trajectories over the stance phase. For each joint, trajectories were time-normalized to 0--100\% of the stance phase and averaged across steady-state cycles before computing $r$. Higher $r$ indicates stronger left--right coordination.
For each condition, stance-time and joint-level symmetry metrics were averaged across steady-state gait cycles for analysis.

\subsection{Simulation, Training, and Evaluation Setup}

All experiments were conducted in simulation using the MuJoCo physics engine (version 3.4.0) within the MyoAssist reinforcement learning framework. We used a 3D musculoskeletal model with 26 lower-limb muscles that generated joint torques through muscle activation. The modeled ankle exoskeleton is a simplified, custom device in the MyoAssist model, consisting of a rigid shank cuff and foot attachment connected by a powered actuator. The actuator applies assistive torque about the ankle joint (single-DoF actuation; Exo\_R in the XML), and the exoskeleton mass (1.165~kg) is attached to the distal segment of the right shank in simulation.

All experiments were performed on flat terrain under identical simulation settings with a nominal reference gait speed of 1.25~m/s; the resulting steady-state speed was allowed to vary across conditions. Control commands from the policies were applied at 30~Hz, while the physics simulation was integrated at 1200~Hz to ensure numerical stability and accurate contact dynamics. Imitation-based training used a reference gait trajectory defined by joint kinematics and end-effector (foot and toe) trajectories. Training was conducted with 32 parallel simulation environments to improve sample efficiency and reduce variance across rollouts.
%All simulations were executed offline without real-time constraints.

Training was performed using the Proximal Policy Optimization (PPO) algorithm \cite{schulman2017proximal}. In Stage~1, a baseline human locomotion policy was first trained on the non-impaired musculoskeletal model (without an exoskeleton) using imitation-based rewards. This baseline policy produced stable, periodic walking behavior under nominal conditions and served as the initialization for subsequent training.

In Stage~2, we first performed transfer learning from Stage~1 baseline under progressive unilateral muscle weakness by modifying the musculoskeletal model to reduce maximum muscle strength in the right leg. The baseline policy was used to initialize training for each impaired condition, resulting in stable but asymmetric gait patterns under different weakness levels. For the $50\%$ strength condition ($\alpha=0.5$), we further conducted human--exoskeleton joint training in a centralized-critic, decentralized-actor setting. A shared centralized critic estimated the value function based on joint observations, while each agent maintained its own actor network. Both policies were updated concurrently using the same advantage estimates, and the human policy continued to adapt to the evolving exoskeleton behavior during training. Unless otherwise specified, all training runs used identical reinforcement learning parameters across conditions. PPO training employed a discount factor $\gamma = 0.99$, a generalized advantage estimation parameter $\lambda = 0.95$, a learning rate of $5\times10^{-5}$, and a clipping range of 0.15. Each policy update used 2048 environment steps per rollout and was optimized for 20 epochs with a batch size of 16384 \cite{tan2025myoassist}. We observed that the Stage~1 baseline policy generally converged around $2.5\times10^{8}$ time steps, while Stage~2 training under asymmetric impairment conditions required roughly $0.3\times10^{8}$ additional time steps to reach comparable stability. Policies were trained for a total of $4\times10^{8}$ simulation time steps.

For evaluation, we used normalized foot/toe contact signals with thresholds of 0.05 (heel strike/toe-off detection) and 0.1 (stance ratio), both dimensionless. A gait cycle was defined as the interval from one heel strike to the subsequent heel strike of the same leg. The stance phase was defined by the heel-strike and toe-off events, and stance duration was computed as the fraction of the gait cycle between these two events for each leg.

Evaluation focused on steady-state walking. For each condition, gait cycles were selected from the final portion of the simulation after walking behavior had stabilized. Steps separated by less than 300~ms were excluded to avoid spurious detections. Joint-level symmetry metrics defined in Section~\ref{sec:metrics} were computed during the stance phase using joint angle trajectories. For each joint (hip, knee, and ankle), stance-phase ROM was computed to obtain the ROM-based SI, and trajectory correlation $r$ was computed between left and right joint angles. Metrics were then averaged across steady-state gait cycles.

For kinematic visualization, joint angle trajectories for the left and right legs were analyzed separately. Gait cycles were segmented independently for each leg and time-normalized to 0-100\% of the gait cycle using linear interpolation to 100 sample points. Mean joint-angle trajectories and standard deviations were computed across all analyzed cycles to quantify inter-cycle variability.

\section{Results}

\subsection{Gait Asymmetry under Unilateral Muscle Weakness}

Figure~2 shows hip, knee, and ankle trajectories under unilateral right-leg weakness.
%For each impairment level, training was initialized from the normal-walking baseline and then continued under the corresponding reduced muscle strength. 
As right-leg muscle strength decreased (100\%, 75\%, 50\%, and 25\%), gait asymmetry increased consistently across temporal, kinetic, and kinematic measures. Mean steady-state walking speed also decreased with weakness (1.07, 1.05, 0.99, and 0.85~m/s for $\alpha=1.0, 0.75, 0.5, 0.25$, respectively).

Under the 100\% strength condition, coordination between the two limbs was high, with strong joint-trajectory correlations (ankle $r=0.974$, knee $r=0.988$, hip $r=0.993$; Table~\ref{tab:panel_summary}). As weakness increased, the ankle exhibited the most consistent degradation in symmetry. In terms of ROM SI, ankle SI moved farther away from zero in magnitude, from $+6.4\%$ at 100\% strength to $-12.9\%$ at 75\%, $-25.8\%$ at 50\%, and $-47.1\%$ at 25\% strength, indicating progressively larger inter-limb differences in ankle motion. Because toe-off timing within a gait cycle corresponds to the stance-phase fraction for that limb, we report the toe-off timing gap for intuitive visualization, while $\text{Asym}_{\text{stance}}$ (Eq.~\ref{eq:stance_asym}) provides the normalized summary metric.

Temporal asymmetry increased with weakness, with the inter-limb toe-off gap widening from $\sim$2\% at 100\% strength to $\sim$7--8\% under moderate-to-severe weakness (Table~\ref{tab:panel_summary}).

\begin{table*}[t]
\centering
\caption{Summary metrics under each condition. SI: ROM-based symmetry index computed using Eq.~(\ref{eq:SI}); $r$: Pearson correlation between left and right joint angle trajectories during stance. Toe-off is reported as the percentage of the gait cycle.}
\label{tab:panel_summary}
\small
\setlength{\tabcolsep}{4pt}
\renewcommand{\arraystretch}{1.12}
\begin{tabular}{l c c c c c}
\hline
Condition & Peak Force (L/R) [N] & Toe-off (L/R) [\%] &
Ankle (SI, $r$) & Knee (SI, $r$) & Hip (SI, $r$) \\
\hline
100\% strength ($\alpha=1.0$) &
$1319.1\pm90.3$ / $1422.4\pm89.3$ &
57 / 55 &
$+6.4$, $0.974$ &
$-9.7$, $0.988$ &
$-4.9$, $0.993$ \\
75\% strength ($\alpha=0.75$) &
$1755.7\pm112.1$ / $1165.4\pm78.6$ &
62 / 55 &
$-12.9$, $0.961$ &
$-19.8$, $0.957$ &
$+0.2$, $0.996$ \\
50\% strength ($\alpha=0.5$) &
$1681.6\pm130.5$ / $1053.3\pm90.5$ &
60 / 53 &
$-25.8$, $0.948$ &
$+22.7$, $0.914$ &
$-7.0$, $0.992$ \\
25\% strength ($\alpha=0.25$) &
$1688.8\pm93.3$ / $1002.0\pm88.4$ &
61 / 53 &
$-47.1$, $0.889$ &
$+29.6$, $0.841$ &
$+7.1$, $0.949$ \\
50\% strength + Exo ($\alpha=0.5$) &
$1686.3\pm117.0$ / $878.6\pm132.8$ &
60 / 54 &
$-18.5$, $0.966$ &
$-12.0$, $0.914$ &
$+4.1$, $0.977$ \\
\hline
\end{tabular}
\end{table*}

\begin{figure}[t]
    \centering
    \includegraphics[width=1.05\linewidth]{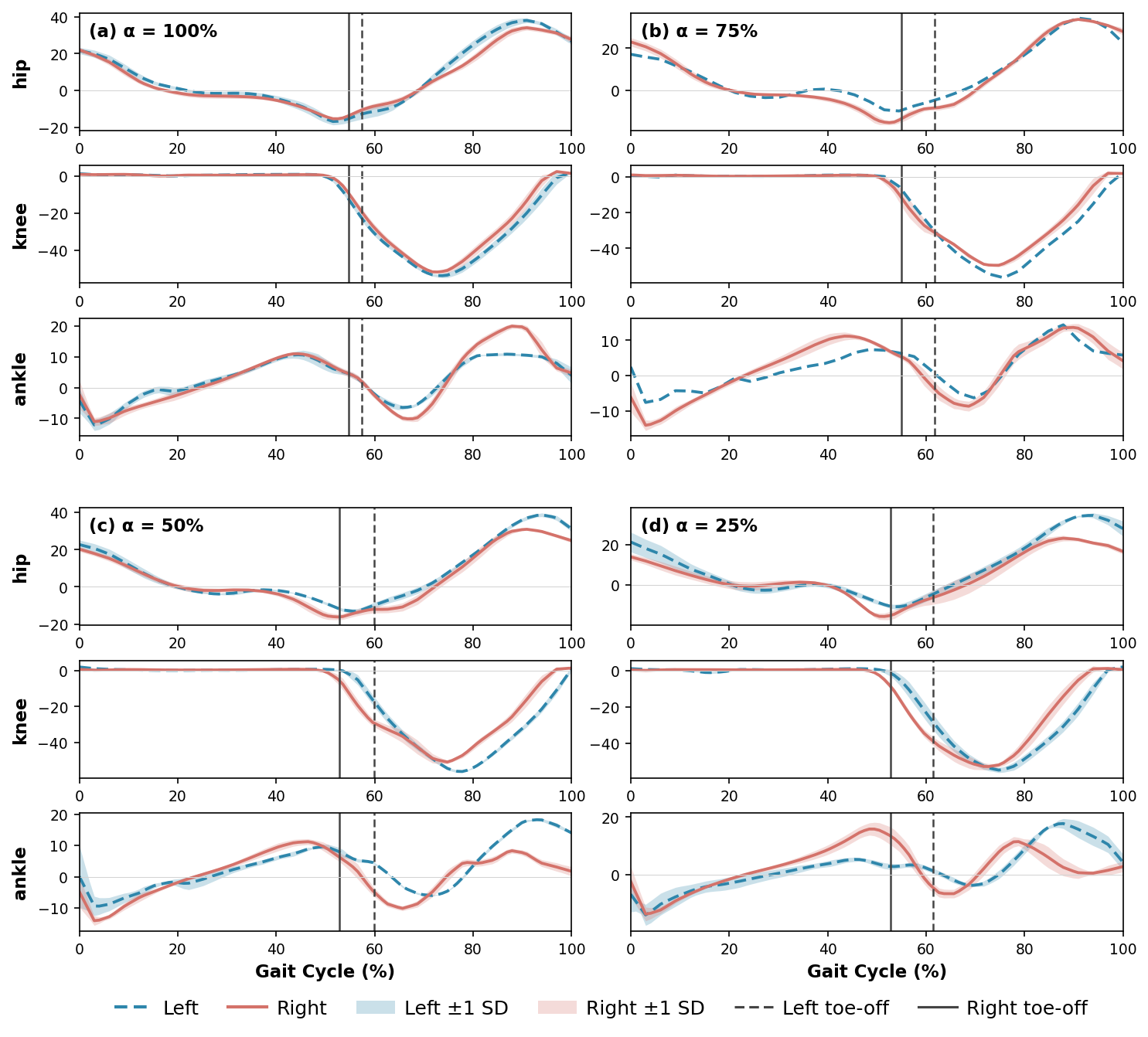}
    \caption{
    Joint angle trajectories during steady-state walking under progressive unilateral right-leg muscle weakness.
    Panels correspond to different muscle strength levels:
    (a) 100\% strength ($\alpha = 1.00$),
    (b) 75\% strength ($\alpha = 0.75$),
    (c) 50\% strength ($\alpha = 0.50$),
    and (d) 25\% strength ($\alpha = 0.25$).
    Solid and dashed lines denote right and left legs, respectively; shaded regions indicate $\pm$1 SD across gait cycles. Toe-off is marked by vertical lines (solid right, dashed left).
    }
    \label{fig:joint_kinematics}
\end{figure}

\subsection{Asymmetry in Contact Forces and Joint Kinematics}
\begin{figure}[t]
    \centering
    \includegraphics[width=1.0\linewidth]{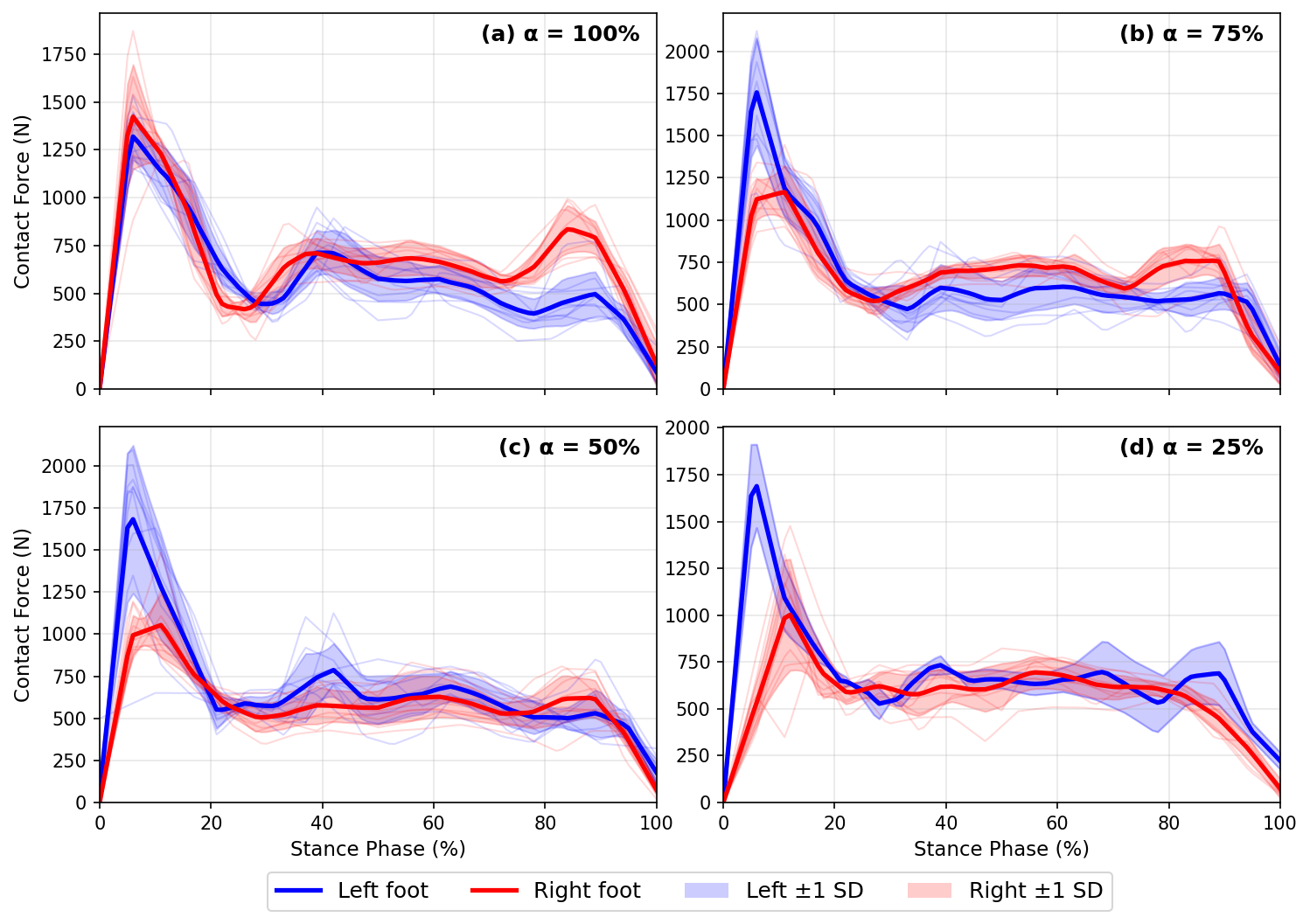}
    \caption{
    Ground contact force profiles during stance under progressive unilateral right-leg muscle weakness (see panel labels for $\alpha$). 
    Blue/red: left/right foot; shaded: $\pm$1 SD across gait cycles.
    }
    \label{fig:stance_force}
\end{figure}
As expected, peak contact forces showed a clear redistribution toward the unimpaired limb as muscle strength decreased. At 75\% strength, the left peak force increased to $1755.7\pm112.1$\,N while the right decreased to $1165.4\pm78.6$\,N, and this left-dominant loading pattern persisted at 50\% and 25\% strength (Table~\ref{tab:panel_summary}). 

Kinematic asymmetry was most pronounced at distal joints. In addition to the monotonic increase in ankle SI magnitude, ankle correlation decreased from $r=0.974$ at 100\% strength to $r=0.889$ at 25\% strength. Knee coordination also deteriorated under severe weakness (knee $r=0.841$ at 25\% strength), whereas hip kinematics remained relatively preserved (hip correlation remained $\ge 0.949$ across all conditions). Overall, unilateral weakness primarily disrupted distal inter-limb coordination during stance, accompanied by compensatory load redistribution on the unimpaired limb. 

\subsection{Ankle Exoskeleton Assistance at 50\% Strength}

Under the 50\% strength condition ($\alpha=0.5$), ankle exoskeleton assistance improved kinematic symmetry relative to the unassisted impaired gait. Ankle ROM SI decreased in magnitude from $-25.8\%$ to $-18.5\%$, and ankle correlation increased from $r=0.948$ to $r=0.966$ (Table~\ref{tab:panel_summary}). Knee symmetry also improved, with knee SI changing from $+22.7\%$ to $-12.0\%$.

\begin{figure}[t]
    \centering
    \includegraphics[width=0.95\linewidth]{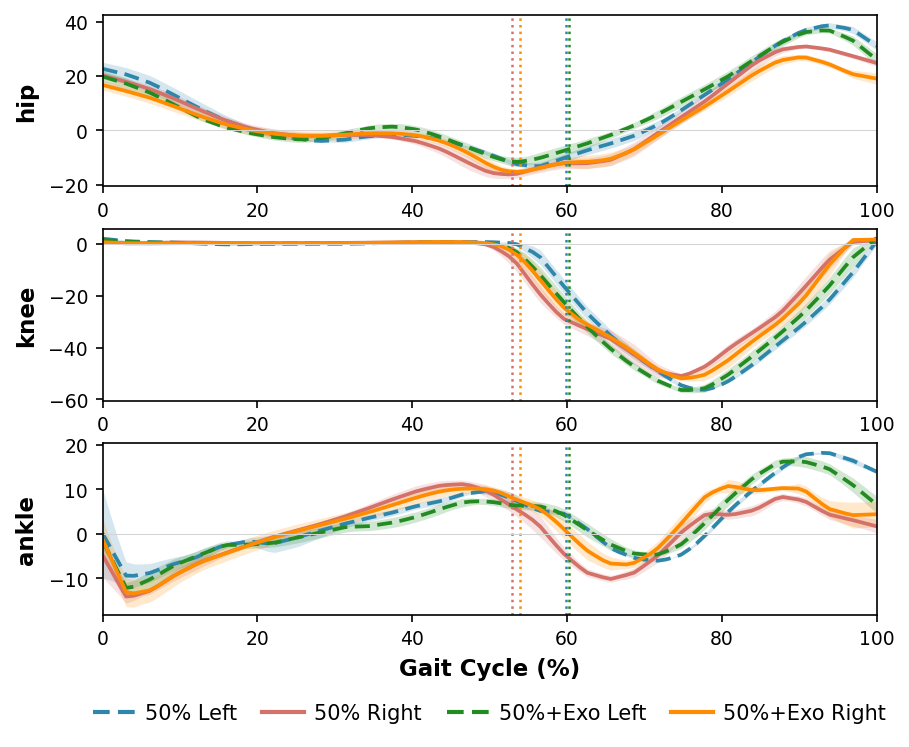}
    \caption{
    Joint angle trajectories at 50\% strength ($\alpha=0.5$) with and without ankle exoskeleton assistance. ``50\%'' denotes unassisted and ``50\%+Exo'' denotes assisted. Solid/dashed lines indicate right/left legs; shaded regions show $\pm1$ SD; vertical lines mark toe-off.
    }
    \label{fig:exo_50_compare}
\end{figure}

Figure~\ref{fig:exo_50_compare} compares joint angle trajectories at 50\% strength with and without ankle assistance. The exoskeleton condition shows closer left--right alignment in ankle kinematics and modest improvements in knee coordination, while hip trajectories remain broadly similar. Toe-off timing also becomes slightly more aligned across limbs under assistance.

\begin{figure}[t]
    \centering
    \includegraphics[width=1.0\linewidth]{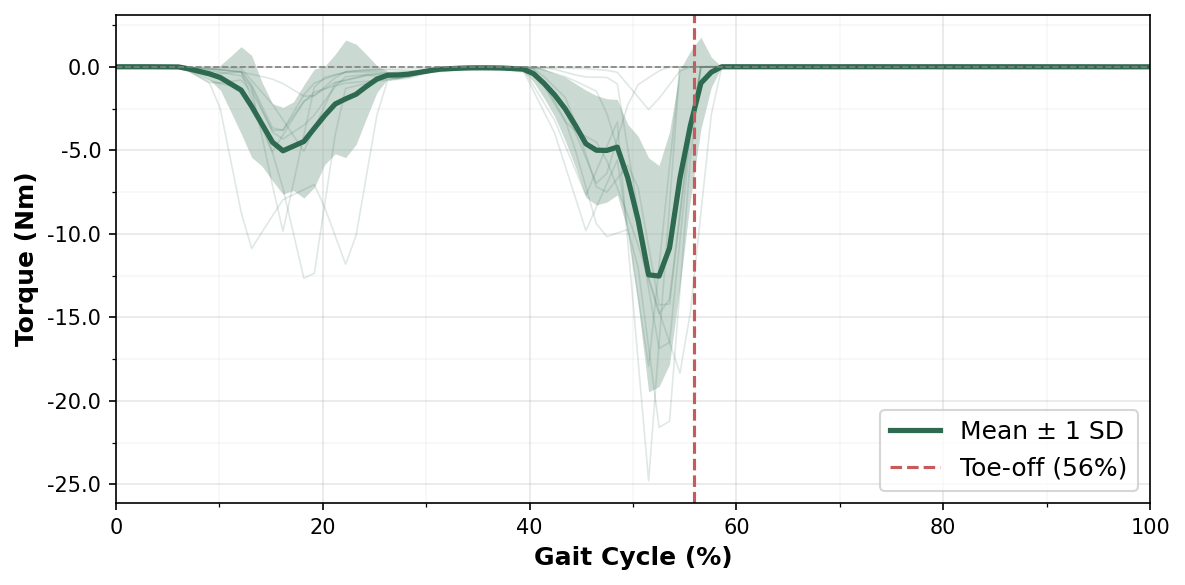}
    \caption{
    Learned right-leg ankle exoskeleton torque profile at 50\% muscle strength ($\alpha=0.5$).
    The curve shows the mean commanded ankle torque over one gait cycle for the assisted condition (50\%+Exo), with shaded regions indicating $\pm 1$ SD across gait cycles.
    Vertical line marks right-leg toe-off timing.
    Positive torque denotes dorsiflexion and negative torque denotes plantarflexion.
    }
    \label{fig:exo_torque_50}
\end{figure}

Figure~\ref{fig:exo_torque_50} shows the mean learned right-leg ankle exoskeleton torque profile over one gait cycle at 50\% strength. The assistance is phase-dependent, with a larger plantarflexion torque during late stance corresponding to push-off.

Temporal symmetry improved modestly as well. The right side toe-off gap decreased from 7\% without exoskeleton (60\% vs 53\%) to 6\% with exoskeleton (60\% vs 54\%). In contrast, peak contact force remained left-dominant under assistance, and the right peak force decreased from $1053.3\pm90.5$\,N to $878.6\pm132.8$\,N. These results indicate that ankle assistance improved inter-limb kinematic coordination and timing, but did not fully restore symmetric loading at this impairment level.

\section{Discussion}

In this study, we employed a reinforcement learning–based musculoskeletal simulation framework to examine how unilateral muscle weakness affects gait symmetry and how targeted ankle assistance influences asymmetric walking behavior. 

The results indicate that unilateral muscle weakness alone is sufficient to produce both temporal and kinematic gait asymmetry, even though the controller was not given any explicit symmetry-related objective. As muscle strength decreased, stance duration progressively shifted toward the unimpaired limb, and coordination between left and right joint motions gradually deteriorated. These changes are consistent with compensatory strategies commonly observed in impaired gait, in which load is redistributed to maintain stability and forward progression. Such asymmetry patterns closely mirror gait characteristics commonly reported in individuals with hemiplegia following stroke, including prolonged stance on the unimpaired limb, reduced inter-limb coordination, and compensatory load redistribution\cite{patel2011successful}. 

Applying ankle assistance to the impaired limb reduced gait asymmetry across multiple measures. The learned assistance was phase-dependent (Fig.~\ref{fig:exo_torque_50}), with torque concentrated in early stance and terminal stance and minimal assistance during mid-stance, which is consistent with the higher ankle moment demand during loading response and push-off. Even without an explicit symmetry objective, augmenting weak-side push-off and stance control can shift toe-off timing and ankle kinematics toward the nominal pattern, improving symmetry; however, the profile is likely task-optimal for imitation tracking rather than explicitly optimized for symmetry restoration. Improvements were observed in toe-off timing and in joint-level coordination during stance. Although assistance was only at the ankle joint, its effects were not limited to local ankle motion but extended to more proximal joints, indicating that distal assistance can influence whole-limb gait dynamics. These results suggest that ankle-level intervention may be sufficient to partially compensate for unilateral muscle weakness, particularly with respect to stance-phase kinematic coordination and temporal symmetry.

However, ankle assistance did not improve ground contact force loading symmetry. Although kinematic symmetry improved, peak ground contact forces remained biased toward the unimpaired limb. This indicates that ankle assistance mainly affected distal joint kinematics and temporal timing, but was insufficient to rebalance inter-limb loading. One possible explanation is that ground contact forces depend more on proximal joint coordination and whole-body load distribution, including hip and knee mechanics and center-of-mass control. As a result, ankle-only assistance may be effective at improving inter-limb coordination, but may not be sufficient to fully restore load symmetry under moderate levels of  impairment (e.g., moderate muscle weakness).

\subsection{Limitations and Future Work}
This study has several limitations. First, all experiments were conducted exclusively in simulation and were not validated in human participants. While simulation enables precise control over muscle strength and systematic testing of assistive strategies, experimental studies will be necessary to confirm real-world effectiveness and safety. Second, weakness was applied uniformly to all right-limb muscles; future work will test muscle-specific and subject-specific weakness profiles. Third, the analysis focused on level-ground walking with a nominal reference gait speed (1.25~m/s), and did not include environmental variability, task transitions, or external perturbations. Evaluating walking across a broader range of  speeds, inclines, or under perturbations would help assess the robustness and generalizability of the learned assistance strategies. Finally, only ankle-level assistance was considered. While it improved distal kinematics and temporal coordination, loading asymmetry persisted, indicating that multi-joint or whole-body assistance strategies may be required to achieve more fully symmetric gait patterns.

\section{Conclusion}

This study leveraged a reinforcement learning–based musculoskeletal simulation to examine gait asymmetry caused by unilateral muscle weakness and to evaluate the effects of targeted ankle assistance. Progressive weakness resulted in increasingly pronounced temporal and kinematic asymmetries, particularly at the ankle, reflecting disrupted inter-limb coordination and altered stance timing. Ankle assistance at 50\% strength partially improved inter-limb kinematic coordination and toe-off timing, demonstrating that targeted support can mitigate some functional deficits. However, peak contact forces remained biased toward the unimpaired limb, indicating that ankle-only assistance is insufficient to fully restore loading symmetry. These findings have important implications for the design of assistive devices: while isolated ankle support can improve certain kinematic aspects of gait, achieving comprehensive gait restoration may require multi-joint or adaptive control strategies. The simulation-based approach provides a platform for exploring the severity of the impairment and assistive strategies, supporting future validation in human experiments.

\bibliographystyle{unsrt}
\bibliography{references}

\end{document}